\pdfoutput=1

\documentclass[11pt]{article}

\usepackage[final]{acl}

\usepackage{times}
\usepackage{latexsym}
\usepackage{booktabs}
\usepackage{xspace}
\usepackage{subfig}

\usepackage[T1]{fontenc}

\usepackage[utf8]{inputenc}

\usepackage{microtype}

\usepackage{inconsolata}

\usepackage{graphicx}

\usepackage{amsmath}
\usepackage{amssymb}
\usepackage{booktabs}
\usepackage{color}
\usepackage{colortbl}
\usepackage{graphicx}
\usepackage{makecell}
\usepackage{mathtools}
\usepackage{multirow}
\usepackage{xltabular}
\usepackage{xspace}

\usepackage{ulem}
\usepackage{xcolor}

\definecolor{yellowhl}{RGB}{255, 255, 153} 
\definecolor{greenhl}{RGB}{180, 238, 180}  
\definecolor{bluehl}{RGB}{173, 216, 230}   
\newcommand{\modelname}{IRAKE\xspace}

\title{Avoiding Knowledge Edit Skipping in Multi-hop Question Answering with Guided Decomposition}

\author{
    Yi Liu$^\dagger$ \quad
    Xiangrong Zhu$^\dagger$ \quad
    Xiangyu Liu$^\dagger$ \quad
    Wei Wei$^\dagger$ \quad
    Wei Hu$^{\dagger,\,\ddagger,\,}$\thanks{Corresponding author} \\
    $^\dagger$ State Key Laboratory for Novel Software Technology, Nanjing University, China \\
    $^\ddagger$ National Institute of Healthcare Data Science, Nanjing University, China \\
    \texttt{\{yiliu07, xrzhu, xyl, weiw\}.nju@gmail.com, whu@nju.edu.cn} 
}

\begin{document}
\maketitle
\begin{abstract}
In a rapidly evolving world where information updates swiftly, knowledge in large language models (LLMs) becomes outdated quickly. Retraining LLMs is not a cost-effective option, making knowledge editing (KE) without modifying parameters particularly necessary. 
We find that although existing retrieval-augmented generation (RAG)-based KE methods excel at editing simple knowledge, they struggle with KE in multi-hop question answering due to the issue of ``edit skipping'', which refers to skipping the relevant edited fact in inference. In addition to the diversity of natural language expressions of knowledge, edit skipping also arises from the mismatch between the granularity of LLMs in problem-solving and the facts in the edited memory.
To address this issue, we propose a novel \textbf{I}terative \textbf{R}etrieval-\textbf{A}ugmented \textbf{K}nowledge \textbf{E}diting method with guided decomposition (\modelname) through the guidance from single edited facts and entire edited cases. 
Experimental results demonstrate that \modelname mitigates the failure of editing caused by edit skipping and outperforms state-of-the-art methods for KE in multi-hop question answering.

\end{abstract}

\section{Introduction}
\label{sect:intro}

Contemporary large language models (LLMs) have achieved impressive performance comparable to humans in many tasks such as question answering~\citep{kamalloo23llmqa,singhal25llmqa}, writing assistance~\citep{yuan22llmwriting,jakesch23llmwriting}, and code generation~\citep{liu23llmcode,zhang23llmcode}.
As a world model \citep{ha18worldmodel,hao23worldmodel}, the temporal dimension of data is of great significance to LLMs in their processes of understanding, memorization, and reasoning. 
However, current LLMs struggle to adapt more flexibly and cost-effectively to the vast and ever-changing data generated in a rapidly evolving world.
The cost of retraining LLMs for minor updates of data is prohibitively high, which leads to the need for knowledge editing (KE) without modifying model parameters.
A series of methods based on retrieval-augmented generation (RAG) ~\citep{mitchell22serac,zheng23ike} have been developed.
They guide LLMs to generate outputs that meet editing requirements by constructing an edited memory containing factual knowledge of edits and show effectiveness in solving editing tasks for single knowledge pieces.

However, a more challenging task for KE is whether the edited model can correctly solve complex questions whose results should change as the impact of KE. 
This task is referred to as multi-hop question answering for knowledge editing (MQuAKE)~\citep{zhong23mquakemello}.
MQuAKE requires a sequence of interconnected knowledge facts to reach the final answer.
Most of the existing methods adopt a framework of decomposing the complex question first and retrieving edited facts that may be involved from edited memory based on the decomposed subquestions~\citep{zhong23mquakemello, gu24pokemqa, wang24deepedit, lu24kedkg}. 

The KE methods based on edited memory and RAG rely on the textual representation of natural language (as opposed to directly modifying model parameters).
However, diversity prevails in the expression of knowledge in natural language. 
For example, the wife of the president of the United States can be expressed as ``First Lady of the United States'', or ``the spouse of the president of the United States''. 
Moreover, when there is insufficient context, the perspectives for understanding a question and the methods for solving it are highly flexible and uncertain.
For example, for the question ``Who is the successor of Tarja Turunen?'', it can be understood in terms of the successor in her musical genre, or the successor to her position as the lead singer in the band.
Under the combined influence of the above circumstances, LLMs may suffer from the issue of edit skipping, which means skipping the impact of edited knowledge facts when answering some forms of multi-hop questions, thus leading to ineffective editing.
The issue of edit skipping is rooted in the mismatch between the granularity of decomposing complex questions and the granularity of relevant edited knowledge facts in the edited memory.

\begin{figure}[t]
\centering
\includegraphics[width=\columnwidth]{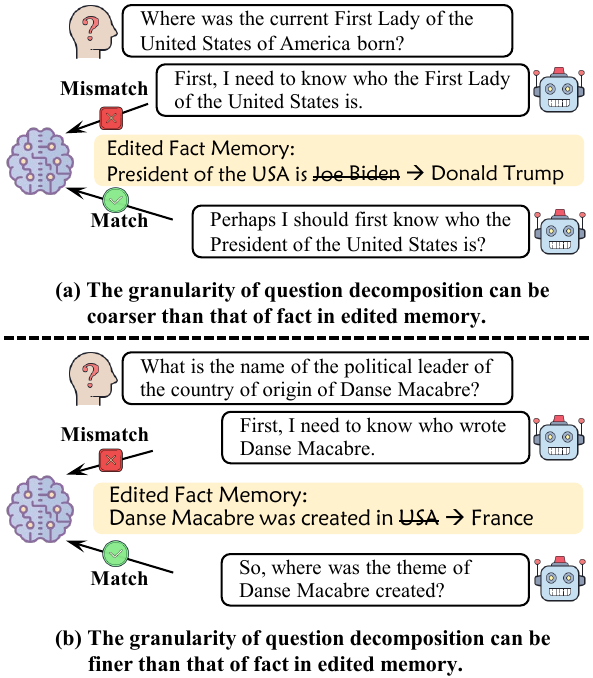}
\caption{The granularity of question decomposition may not match the granularity of relevant edited facts in the edited fact memory.}
\label{fig:motivation}
\end{figure}

As shown in Figure~\ref{fig:motivation}(a), when solving the question ``Where was the current First Lady of the United States of America born?'', the granularity of the subquestion ``Who is the First Lady of the United States'' is coarser than that of the edited fact ``The president of the USA is Donald Trump'', making it challenging to directly align the subquestion with the edited fact. 
If an LLM directly answers this subquestion, it may skip the influence of the edited fact. 
Similarly, in Figure 1(b), the decomposition granularity of the subquestion ``Who wrote Danse Macabre?'' is finer than that of the edited fact ``Danse Macabre was created in France'', which also increases the difficulty of matching the subquestion with the edited knowledge fact and may lead LLMs to skip the relevant edit.

To mitigate the edit skipping issues, we propose an iterative retrieval-augmented KE method, \modelname, based on edit-guided question decomposition, where the guidance lies at both the edited fact level and the edited case level. 
At the edited fact level, \modelname selects the edited fact that is most helpful for the decomposition of the current complex question through pre-retrieval and judgment. 
It uses the corresponding atomic question to guide the question decomposition.
At the edited case level, \modelname searches for the edited case with the most similar question as the current question and uses its solution to construct the dynamic guidance prompt and guide the question decomposition.
We also design a state backtracking mechanism to alleviate the impact of failed guidance.
Extensive experiments demonstrate that \modelname outperforms state-of-the-art KE methods. 

Our main contributions are outlined as follows:
\begin{itemize}
    \item We investigate the problem of edit skipping caused by the knowledge granularity mismatch in MQuAKE and introduce a new ``retrieve-then-decompose'' method \modelname.

    \item We propose decomposition guidance from the perspectives of edited facts and cases. 
    We also design a state backtracking mechanism to alleviate the impact of failed guidance.

    \item We evaluate the effectiveness of \modelname in multi-hop question answering. 
    It outperforms state-of-the-art KE baselines. 
    All the modules that we develop are also effective.
\end{itemize}

\section{Related Work}
\label{sect:related}

\textbf{Knowledge Editing} (KE)~\citep{mazzia23kesurvey,zhang24kesurvey,wang25kesurvey} aims to efficiently modify the memory of an LLM regarding specific knowledge without requiring full retraining, which is often impractical due to high computational costs. 
Existing KE methods can be generally classified into two categories: parameter modification-based methods \citep{meng22rome,meichell22mend,meng23memit} and retrieval-augmented methods \citep{mitchell22serac,zheng23ike,cohen24ice}.
Finetuning all model parameters can lead to performance degradation due to overfitting on limited edited knowledge \citep{hu22lora,ding23finetune} .
Parameter modification-based KE methods \citep{meng22rome,meichell22mend,meng23memit} identify and update only the parameters relevant to specific edits, keeping the rest of the model frozen.
Retrieval-augmented KE methods \citep{mitchell22serac,zheng23ike,cohen24ice} maintain an editable knowledge base and retrieve relevant edits at inference time to enrich the input and suppress outdated information, leveraging the prompt-based reasoning ability of LLMs.
While parameter modification-based KE methods struggle with the interpretability \citep{hase23localization} of why specific knowledge is tied to particular parameters, retrieval-augmented KE methods offer a balance between accuracy and efficiency.

\medskip\noindent\textbf{Knowledge Editing in Multi-hop Question Answering} is challenging, as it requires reasoning over and linking multiple facts to derive the answer.
Previous methods~\citep{zhong23mquakemello,gu24pokemqa,wang24deepedit,lu24kedkg,wang24editcot,cikmshi} address KE in multi-hop question answering by decomposing complex questions into subquestions and resolving them iteratively to obtain the final answer.
However, most of them adopt a decomposition-then-retrieval paradigm, without considering the issue of mismatch between decomposition and target edits, which may fail to retrieve the necessary edits and result in the issue of edit skipping.
To address this issue, \modelname performs a pre-retrieval step before decomposition, retrieving relevant candidate edits and past decomposition records to guide subquestion generation.

\begin{figure*}
\centering
\includegraphics[width=\linewidth]{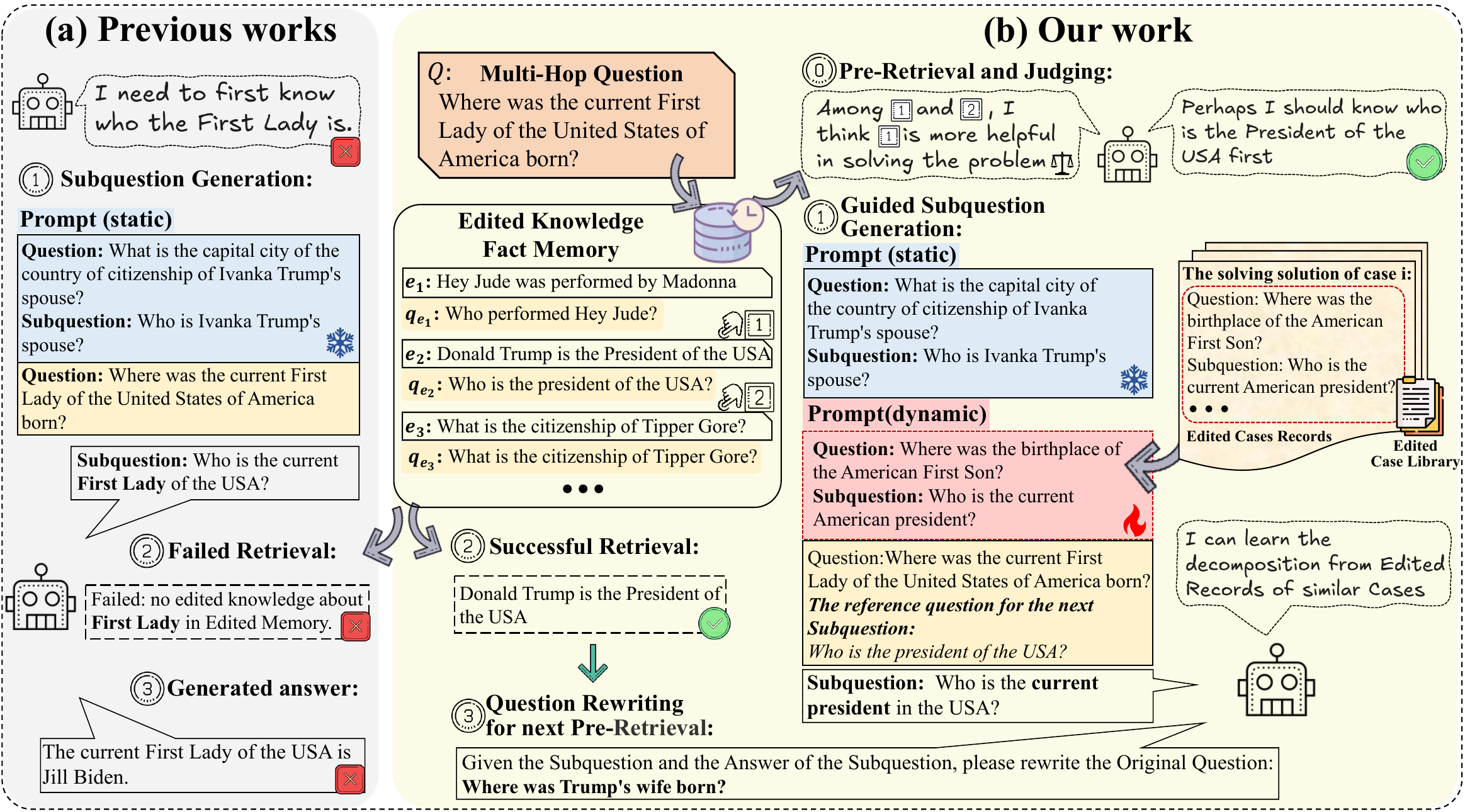}
\caption{Pipeline of the first iteration diagram for solving the multi-hop question $Q$. Part (a) illustrates the commonly used process in previous works, which involves decomposing the question first and then retrieving the edited fact memory. Part (b) is our proposed \modelname, which adopts a process of pre-retrieval first, followed by guided question decomposition, and last fine-grained retrieval.}
\label{fig:framework}
\end{figure*}

\section{Preliminaries}
\paragraph{Notations.}
Following previous works~\citep{zhong23mquakemello, gu24pokemqa, lu24kedkg}, we denote each piece of fact as a triplet $(s,r,o)$, where $s$, $o$, and $r$ represent the subject, object, and relationship, respectively. One factual edit $e$ is modeled as updating the object in the triplet from $o$ to $o^{*}$, denoted by $e = (s,r,o \to o^{*}) $. Meanwhile, the question corresponding to $(s, r)$ in one factual edit $e$ is called the atomic question, denoted by $q_{e}$~\citep{gu24pokemqa}.
In the task of KE, there are usually multiple factual edits, which are stored in the edited fact memory: $\mathcal{E} = \{e_{1}, e_{1},\dots,e_{n}\}$. The relevant edited facts in $\mathcal{E}$ are retrieved and used to modify the knowledge in the LLM.
Without loss of generality, the edited fact memory stores both factual edits and their corresponding atomic questions:
$\mathcal{E}_q = \{(e,q_e)\,|\,e\in \mathcal{E}\}$.

\paragraph{MQuAKE.}
Given a multi-hop question $Q$, answering $Q$ requires sequentially querying and retrieving multiple facts. 
According to the retrieval order of the queries and their corresponding answers, these factual answers can form a chain of facts: $\mathcal{C} = [(s_{1}, r_{1}, o_{1}),\dots,(s_{n}, r_{n}, o_{n})]$, where $o_{i} = s_{i+1}$ and $o_{n}$ is the final answer of $Q$. 
Replacing any fact $(s_{i}, r_{i}, o_{i})$ on this chain with an edited fact $(s_{i}, r_{i}, o_{i}^{*})$ may affect the entire subsequent fact chain: $\mathcal{C}^{*} = [(s_{1}, r_{1}, o_{1}),\dots,(s_{i}, r_{i}, o_{i}^{*}),\dots,$ $(s_{n}^{*}, r_{n}, o_{n}^{*})]$, where $o_{n}^{*}$ is the updated final answer of $Q$.
A multi-hop question and its corresponding edited facts constitute an edited case with one question.
The task of MQuAKE can be formalized as follows: Given an edited fact memory $\mathcal{E}$ and an LLM $M$, derive a conditionally edited language model $M^{*}$. 
For each multi-hop question affected by $\mathcal{E}$, $M^{*}$ should produce the correct answer (successfully completing the edited case). 
The reasoning path in this process needs to align with $\mathcal{C}^{*}$, where $\mathcal{C}^{*}$ denotes the gold path for question $Q$.

\section{Methodology}
\label{sect:method}

\subsection{Workflow of \modelname}
\label{sect:workflow}
Some edited facts that influence complex multi-hop questions can only be pinpointed after the questions are decomposed.
Consequently, most existing works adopt a paradigm of decomposing the question first and then retrieving the edited fact memory.
To address the issue of skipping edited facts due to question decomposition deviation, \modelname adopts a paradigm that involves pre-retrieval, guided question decomposition, and subsequently fine-grained retrieval. 
Technically, our approach mainly encompasses question decomposition guided by the edited fact (cf. Section~\ref{sect:pre_retrieval}), guided question decomposition utilizing dynamic prompts derived from similar edited cases (cf. Section~\ref{sect:dynamic_prompt}), and a state backtracking mechanism to reduce the impact of ineffective guidance paths (cf. Section~\ref{sect:backtracking}).

The specific workflow for resolving the first-round subquestion in the multi-hop question $Q$ is shown in Figure~\ref{fig:framework}. \modelname identifies the edited fact that is effective for problem decomposition through pre-retrieval. The atomic question corresponding to this edited fact is used to guide the question decomposition. Similar edited case records are also retrieved to strengthen the guidance for question decomposition. Then, the decomposed subquestions are used for precise retrieval to retrieve the corresponding answer. Subsequently, \modelname proceeds to resolve the next round of subquestions until the final answer is derived.

\subsection{Decomposition Guided by Edited Facts}
\label{sect:pre_retrieval}
\textit{To provide accurate answers, people often rely on relevant contexts.}
When factual modifications are expected, it is essential to refer to the potentially relevant edited facts to address complex problems effectively.
\modelname is designed to leverage atomic questions corresponding to these edited facts to guide the solution of complex question $Q$.
Specifically, we first employ a pre-retrieval process. It takes $Q$ as input, computes the similarity between $Q$ and all edited facts $e \in \mathcal{E}_q$, and returns the top $n$ edited facts with the highest similarity scores (forming $\mathcal{F}\subseteq\mathcal{E}_q$).
In contrast to the subsequent precise retrieval process which employs high-precision techniques such as thresholding and re-ranking, pre-retrieval is a coarse-grained retrieval process that returns a specified number of edited facts.
However, not all facts in $\mathcal{F}$ are necessarily helpful for solving $Q$, and there may not even exist any helpful facts.
Therefore, we rely on the LLM to evaluate and identify the most relevant edited fact $e$ for edited fact-level guidance. If there is no relevant edited fact, decompose the question directly without guidance.

To effectively use the selected edited fact for guidance, we utilize the atomic question $q_{e}$ corresponding to the edited fact rather than the fact itself.
The reason is that the edited facts often contradict the internal knowledge of LLMs. Directly using it as guidance may amplify the doubts of the model during the reasoning process, potentially leading to significant deviations in the reasoning path.
The atomic question is presented to the LLM in the form of a prompt to facilitate the guidance.

Generally, the original question $Q$ contains at least the relevant information for the first step of decomposition. However, it may not include all the details required for subsequent steps in the resolution process. Therefore, during the first decomposition of question $Q$, the pre-retrieval can utilize the original $Q$ to retrieve facts from the edited fact memory $\mathcal{E}$. For later decomposition steps, we employ question rewriting to expose the information required for pre-retrieval. Specifically, \modelname prompts the LLM to rewrite the original $Q$ into $Q^{'}$ based on the subquestion derived from the previous round of decomposition and its corresponding answer for the next pre-retrieval.

\subsection{Decomposition Guided by Edited Cases}
\label{sect:dynamic_prompt}
\textit{Similar complex multi-hop questions often have analogous decomposition structures, and their potential edited points for edited facts tend to overlap.}
Therefore, we posit that the decomposition records of the edited cases, which correspond to similar questions of $Q$ and have been successfully answered under the influence of edited facts, can guide the decomposition of the current question $Q$.
To this end, we build an edited case library $\mathcal{M}$ that stores the records of successfully completed cases. The key of $\mathcal{M}$ is the question $Q_{i}$ in the case and the value is the corresponding solution record, including the decomposition process of $Q_{i}$. 
The construction and subsequent updates of this edited case library are flexible. We simply start to sample a small number of cases from the training set, yielding promising results.

Before solving the question $Q$, we first use $Q$ to search the edited case library and identify the case corresponding to the question most similar to $Q$. Specifically, the question $Q_{*}$ of the target case is obtained as follows: 
\begin{align}
Q_{*} =\underset{Q_{i}\in \mathcal{M},\,\operatorname{sim}(Q, Q_{i})\ge \theta}{\arg \max} \operatorname{sim}(Q, Q_{i}),
\label{eq:DP}
\end{align}
where $\operatorname{sim}(a,b)$ calculates the similarity between $a$ and $b$, and $\theta$ is the threshold.
We then use the complete decomposition record as the dynamic prompt.
The dynamic prompt is combined with a static prompt(shared by all questions), to guide the decomposition of the current question.
The static prompt outlines the overall objective of the task, while the dynamic prompt provides more refined guidance based on specific cases.

\subsection{State Backtracking Mechanism}
\label{sect:backtracking}
\textit{The guidance may also lead the model to skip the facts that need to be edited.}
For example, for the question $Q$: ``What is the name of the political leader of the country of origin of Danse Macabre?'', under the guidance of the question: ``Who wrote Danse Russe?'' (according to the edited fact: ``Writer of Danse Russe is Camille Saint - Sa\"{e}ns''), a subquestion such as ``Who wrote Danse Macabre?'' may be generated (whereas the expected question is ``where was the theme of Danse Macabre created?'' according to target edited fact: ``Danse Macabre was created in France'').

To alleviate this issue, we propose a state backtracking mechanism.
Specifically, whenever the model performs guided question decomposition to generate a subquestion helpful for solving the original question, it stores a non-guided decomposition state on a stack.
If the helpful subquestion indeed retrieves the edited facts during subsequent precise retrieval, we consider it a successful decomposition and clear the stack.
If the stack is not empty when the final answer is generated, this implies that no edited facts were involved since the last stack reset.
In this case, we backtrack (pop up from the stack) to the previously saved non-guided decomposition state and continue the reasoning process.

\section{Experiments and Results}
\label{sect:exp}

\subsection{Experiment Setup}
\paragraph{Datasets.}
We conduct experiments on the benchmark datasets MQuAKE-2002 and MQuAKE-hard \citep{wang24deepedit} derived from MQuAKE \citep{zhong23mquakemello}. MQuAKE-2002 is filtered to exclude instances of which the ground-truth answers are broken by the new knowledge from other instances. MQuAKE-hard is a more challenging subset of MQuAKE by selecting the instances that contain the highest number of edited facts per instance. 
More details are provided in Appendix~\ref{app:Dataset Statistics}.

\paragraph{Evaluation Metrics.}
Following prior work \citep{zhong23mquakemello, gu24pokemqa, wang24deepedit}, we evaluate model performance under three settings: 1-edited, 100-edited, and all-edited, corresponding to batch sizes of 1, 100, and all, respectively. Each batch provides relevant edited facts for retrieval. We report multi-hop accuracy (Acc) and hop-wise answering accuracy (Hop-Acc). Acc measures whether the edited LLM correctly answers multi-hop questions, while Hop-Acc evaluates whether the predicted reasoning path exactly matches the gold path. Each case includes three generated multi-hop questions. Following prior work \citep{zhong23mquakemello, gu24pokemqa, lu24kedkg}, we consider a case correctly answered if any of the three questions is answered correctly.

\begin{table*}[!t]
\centering
{\small
\begin{tabular}{l|cccccc|cccc}
\toprule
\multirow{4}{*}{Methods} & \multicolumn{6}{c|}{MQuAKE-2002} & \multicolumn{4}{c}{MQuAKE-hard} \\
\cmidrule(lr){2-7}\cmidrule(lr){8-11}
& \multicolumn{2}{c}{1-edited} & \multicolumn{2}{c}{100-edited} & \multicolumn{2}{c|}{All-edited} & \multicolumn{2}{c}{1-edited} & \multicolumn{2}{c}{All-edited} \\
\cmidrule(lr){2-3}\cmidrule(lr){4-5}\cmidrule(lr){6-7}\cmidrule(lr){8-9}\cmidrule(lr){10-11}
& Acc & Hop-Acc & Acc & Hop-Acc & Acc & Hop-Acc & Acc & Hop-Acc & Acc & Hop-Acc \\
\midrule
\rowcolor{gray!20} \multicolumn{11}{c}{\textbf{LLaMa-3-8B}} \\
\midrule
FT & 23.30 & - & \ \ 1.97 & - & \ \ 0.80 & - & \ \ 3.10 & - & \ \ 1.16 & - \\
FT$_{\text{CoT}}$ & 27.17 & \ \ 6.10 & \ \ 2.41 & \ \ 0.04 & \ \ 0.99 & \ \ 0.04 & \ \ 3.80 & \ \ 0.00 & \ \ 1.39 & \ \ 0.00 \\
ROME & 12.37 & - & \ \ 2.47 & - & \ \ 2.37 & - & \ \ 3.20 & - & \ \ 1.63 & - \\
ROME$_{\text{CoT}}$ & 15.27 & \ \ 6.47 & \ \ 4.60 & \ \ 0.03 & \ \ 4.53 & \ \ 0.20 & \ \ 4.20 & \ \ 0.00 & \ \ 2.09 & \ \ 0.00 \\
MEMIT & 13.23 & - & \ \ 8.20 & - & \ \ 4.27 & - & \ \ 4.70 & - & \ \ 2.09 & - \\
MEMIT$_{\text{CoT}}$ & 17.97 & \ \ 7.23 & 11.40 & \ \ 3.47 & \ \ 6.30 & \ \ 0.70 & \ \ 5.10 & \ \ 0.00 & \ \ 2.33 & \ \ 0.00 \\
MeLLo & 36.57 & 11.30 & 21.30& 12.07 & 14.33 & \ \ 7.30 & 10.50 & \ \ 3.50 & \ \ 4.60 & \ \ 0.20 \\
DeepEdit& 40.22 & 12.43 & 32.20 & 16.20 & 17.76 & \ \ 9.50 & 14.50 & \ \ 2.40 & \ \ 8.80 & \ \ 0.30 \\
PokeMQA & 48.23 & 33.60 & 39.13 & 29.88 & 36.81 & 24.97 & 33.96 & 19.51 & 27.16 & 17.14 \\
\modelname (ours) & \textbf{65.30} & \textbf{46.30} & \textbf{58.50} & \textbf{48.50} & \textbf{55.24} & \textbf{44.80} & \textbf{52.50} & \textbf{33.00} & \textbf{40.79} & \textbf{35.90} \\
\midrule
\rowcolor{gray!20} \multicolumn{11}{c}{\textbf{DeepSeek-V2-Lite-16B}} \\
\midrule
MeLLo    & 46.40 & 18.70 & 41.75 & 25.25 & 34.86 & 22.27 & 31.10 & \ \ 1.20  & \ \ 6.29 & \ \ 1.39 \\
DeepEdit & 50.00 & 22.40 & 45.92 & 27.83 & 38.10 & 25.61 & 33.20 & 12.10 & 12.43 & \ \ 6.10 \\
PokeMQA  & 51.40 & 32.50 & 50.25 & 36.10 & 44.78 & 33.76 & 35.40 & 18.30 & 29.83 & 23.76\\
\modelname (ours) & \textbf{56.30} & \textbf{37.40} & \textbf{53.50} & \textbf{43.25} & \textbf{50.51} & \textbf{38.81} & \textbf{50.50} & \textbf{21.40} & \textbf{40.09} & \textbf{27.73} \\
\midrule
\rowcolor{gray!20} \multicolumn{11}{c}{\textbf{GPT-4o-Mini}} \\
\midrule
MeLLo & 33.20 & \ \ 8.20 & 22.90 & 11.50 & 17.20 & \ \ 5.40 & \ \ 8.40 & \ \ 1.20 & \ \ 4.70 & \ \ 0.40 \\
DeepEdit& 41.60 & 13.20 & 34.60 & 17.50 & 20.50 & 10.50 & 15.30 & \ \ 1.80 & \ \ 7.20 & \ \ 0.60 \\
PokeMQA & 51.23 & 34.70 & 38.33 & 27.36 & 34.73 & 23.77 & 34.10 & 21.50 & 26.15 & 15.03 \\
\modelname (ours) & \textbf{56.26} & \textbf{42.51} & \textbf{51.50} & \textbf{46.58} & \textbf{46.75} & \textbf{42.11} & \textbf{41.35} & \textbf{24.81} & \textbf{32.13} & \textbf{24.71} \\
\bottomrule
\end{tabular}
}
\caption{Main results. The best scores for each base LLM are marked in \textbf{bold}.
``-'' denotes ``not applicable''.}
\label{tab:main_res}
\end{table*}

\paragraph{Baselines.}
We compare \modelname with the following baselines (method$_{\text{CoT}}$ denotes the method equipped with a chain-of-thought (CoT) prompt):
\begin{itemize}
\item FT/FT$_{\text{CoT}}$, which simply performs gradient descent on the edits to finetune the model.

\item ROME/ROME$_{\text{CoT}}$~\citep{meng22rome}, which first localizes the factual knowledge at a certain layer in the LLM, and then updates the feedforward network.

\item MEMIT/MEMIT$_{\text{CoT}}$~\citep{meng23memit}, which extends ROME to enable editing a large set of facts through updating the feedforward networks in multiple layers.

\item MeLLo~\citep{zhong23mquakemello}, which designs a prompt to alternatively conduct query decomposition and KE by detecting conflicts between the generated answer and edited facts.

\item DeepEdit~\citep{wang24deepedit}, which employs carefully crafted decoding constraints to improve logical coherence and knowledge integration during multi-hop reasoning.

\item PokeMQA~\citep{gu24pokemqa}, which decomposes knowledge-augmented multi-hop questions and interacts with a detached scope classifier to modulate LLMs' behavior.
\end{itemize}

For a fair comparison, we do not include methods that require fine-tuning on key components (e.g., question decomposition in KEDKG~\citep{lu24kedkg}) or those relying on a complete external knowledge base (e.g., RAE~\citep{cikmshi}), as they address different research objectives and are orthogonal to our work. Implementation details are provided in Appendix~\ref{app:inple-details}, and the prompts used by \modelname are listed in Appendix~\ref{app:prompt}.

\begin{table*}[t]
\centering
\resizebox{\textwidth}{!}{
\begin{tabular}{ccc|cccccc|cccc}
\toprule
\multirow{4}{*}{\makecell*[c]{Fact\\Guided}} & \multirow{4}{*}{\makecell*[c]{Case\\Guided}} & \multirow{4}{*}{Backtrack} & \multicolumn{6}{c|}{MQuAKE-2002} & \multicolumn{4}{c}{MQuAKE-hard} \\
\cmidrule(lr){4-9}\cmidrule(lr){10-13}
& & & \multicolumn{2}{c}{1-edited} & \multicolumn{2}{c}{100-edited} & \multicolumn{2}{c|}{All-edited} & \multicolumn{2}{c}{1-edited} & \multicolumn{2}{c}{All-edited} \\
\cmidrule(lr){4-5}\cmidrule(lr){6-7}\cmidrule(lr){8-9}\cmidrule(lr){10-11}\cmidrule(lr){12-13}
& & & Acc & Hop-Acc & Acc & Hop-Acc & Acc & Hop-Acc & Acc & Hop-Acc & Acc & Hop-Acc \\
\midrule
$\times$ & \checkmark & \checkmark &  43.40 & 30.60 & 48.50 & 40.75 & 42.10 & 31.46 & 43.30 & 28.70 & 33.96 & 30.70\\
\checkmark & $\times$ & \checkmark &  61.50 & 40.50 & 55.50 & 45.50 & 51.59 & 37.95 & 45.30 & 25.40 & 37.06 & 32.40\\
\checkmark & \checkmark & $\times$ &  64.50 & 45.50 & 57.20 & 48.30 & 54.12 & 43.20 & 51.30 & 31.30 & 39.30 & 34.10\\
\checkmark & \checkmark & \checkmark & 65.30 & 46.30 & 58.50 & 48.50 & 55.24 & 44.80 & 52.50 & 33.00 & 40.79 & 35.90 \\
\bottomrule
\end{tabular}
}
\caption{Results of ablation study. ``\checkmark'' and ``$\times$'' denote the enabled and disabled modules, respectively.}
\label{tab:ablation_res}
\end{table*}

\subsection{Main Results}
\label{sect:main_res}
The comparison results in terms of Acc and Hop-Acc between \modelname and baselines are shown in Table~\ref{tab:main_res}.
We have the following observations:
(i) \modelname achieves the best performance on both datasets, under all batch sizes, with all alternative base LLMs.
(ii) Retrieval-augmented KE methods (DeepEdit, PokeMQA, and \modelname) significantly outperform parameter modification-based KE methods (FT, ROME, and MEMIT).
(iii) The choice of base LLMs affects editing performance. For instance, MeLLo and DeepEdit achieve better editing accuracy with DeepSeek-V2-Lite-16B than with LLaMa-3-8B or GPT-4o-Mini, while \modelname consistently excels across various base LLMs.
(iv) The MQuAKE-hard dataset is more challenging than MQuAKE-2002 for both the base LLMs and the editing methods, but \modelname exhibits the least performance degradation.
Runtime and token consumption comparisons are shown in Appendix~\ref{app:exp_runtime}.

\subsection{Ablation Study}
We conduct an ablation study to assess the contributions of different modules in \modelname: the edited fact-level guidance, the edited case-level guidance, and the backtracking mechanism. Table~\ref{tab:ablation_res} presents results using LLaMa-3-8B-instruct as the base LLM.
We have the following observations:
(i) Removing each module results in performance degradation, indicating that each module contributes to the overall effectiveness.
(ii) Removing the fact-level guidance causes the most significant performance drop, suggesting that the edited fact-level guidance helps more in generating subquestions and reasoning paths aligned with the target edits.
(iii) Removing the edited case-level guidance causes a noticeable performance drop on the MQuAKE-hard dataset, meaning that its multi-step decomposition guidance is more helpful for complex questions.
(iv) Removing the backtracking mechanism leads to slight performance degradation, indicating that this module indeed helps mitigate issues caused by misleading guidance.

\begin{figure}[t]
\centering
\includegraphics[width=.9\columnwidth]{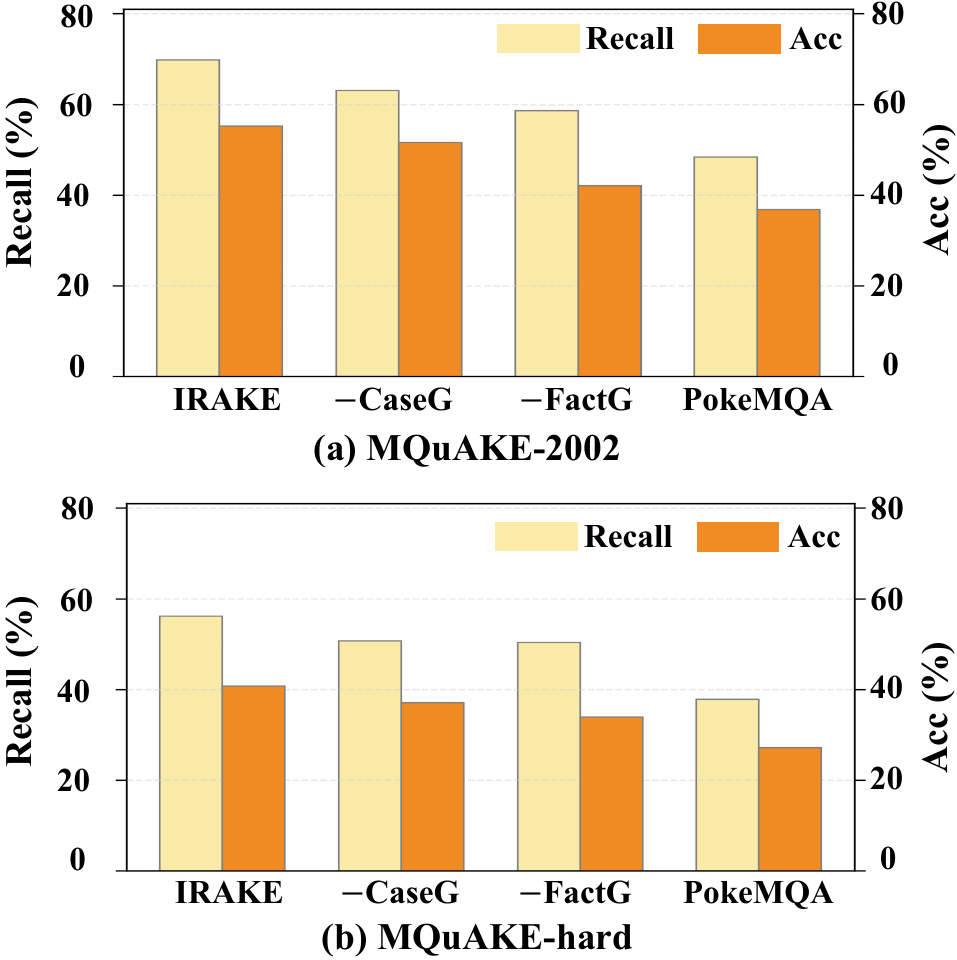}
\caption{Recall of edited facts and Acc of different methods in MQuAKE-2002 and MQuAKE-hard.}
\label{fig:recall_acc_exp}
\end{figure}

\begin{figure}[t]
\centering
\includegraphics[width=.9\columnwidth]{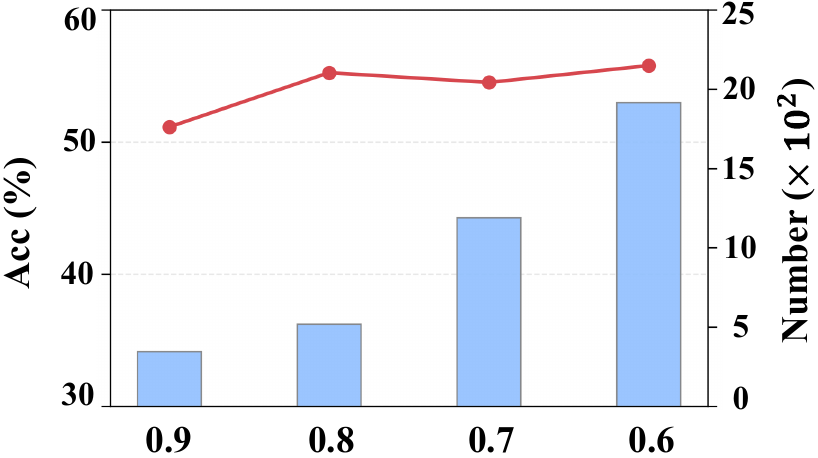}
\caption{Relationship between Acc and the similarity threshold $\theta$ for edited case guidance selection.}
\label{fig:exp3}
\end{figure}

\begin{figure}[t]
\centering
\includegraphics[width=.9\columnwidth]{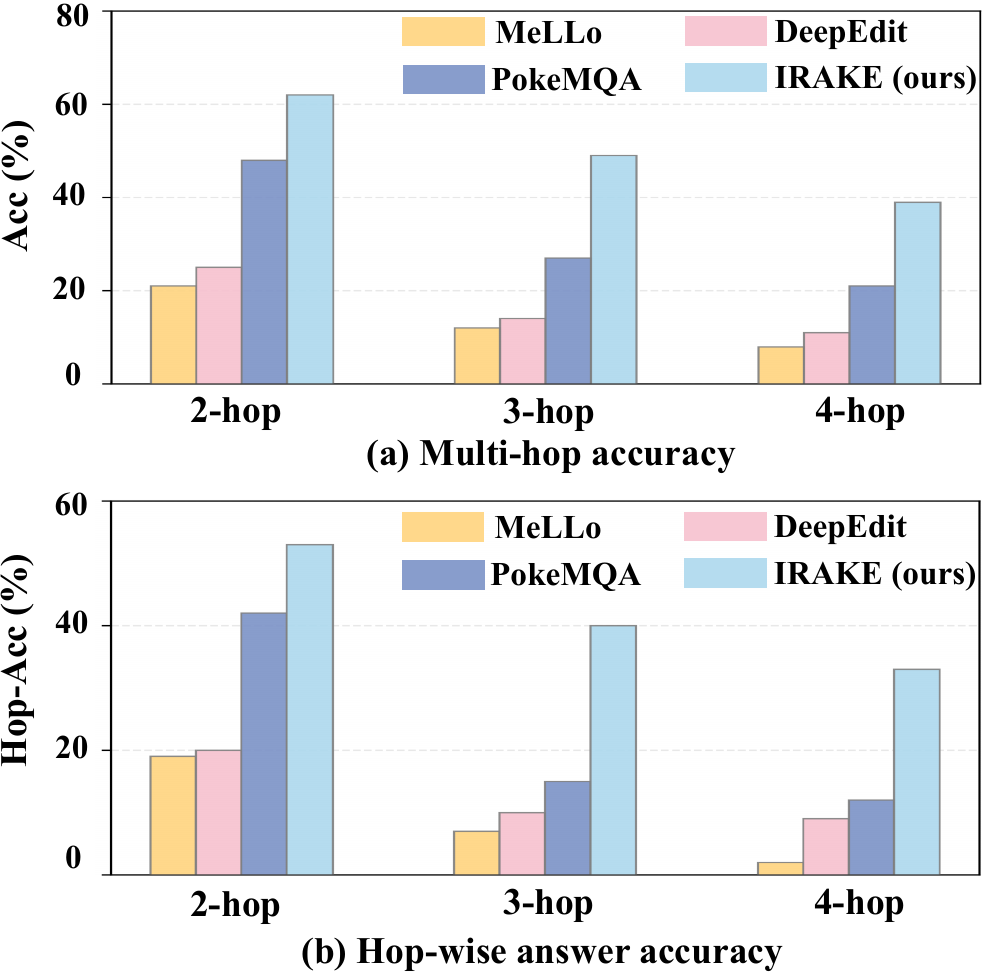}
\caption{Acc and Hop-Acc results on MQuAKE-2002, utilizing different knowledge editing methods.}
\label{fig:multi_hop_exp}
\end{figure}

\begin{figure*}
\centering
\includegraphics[width=\linewidth]{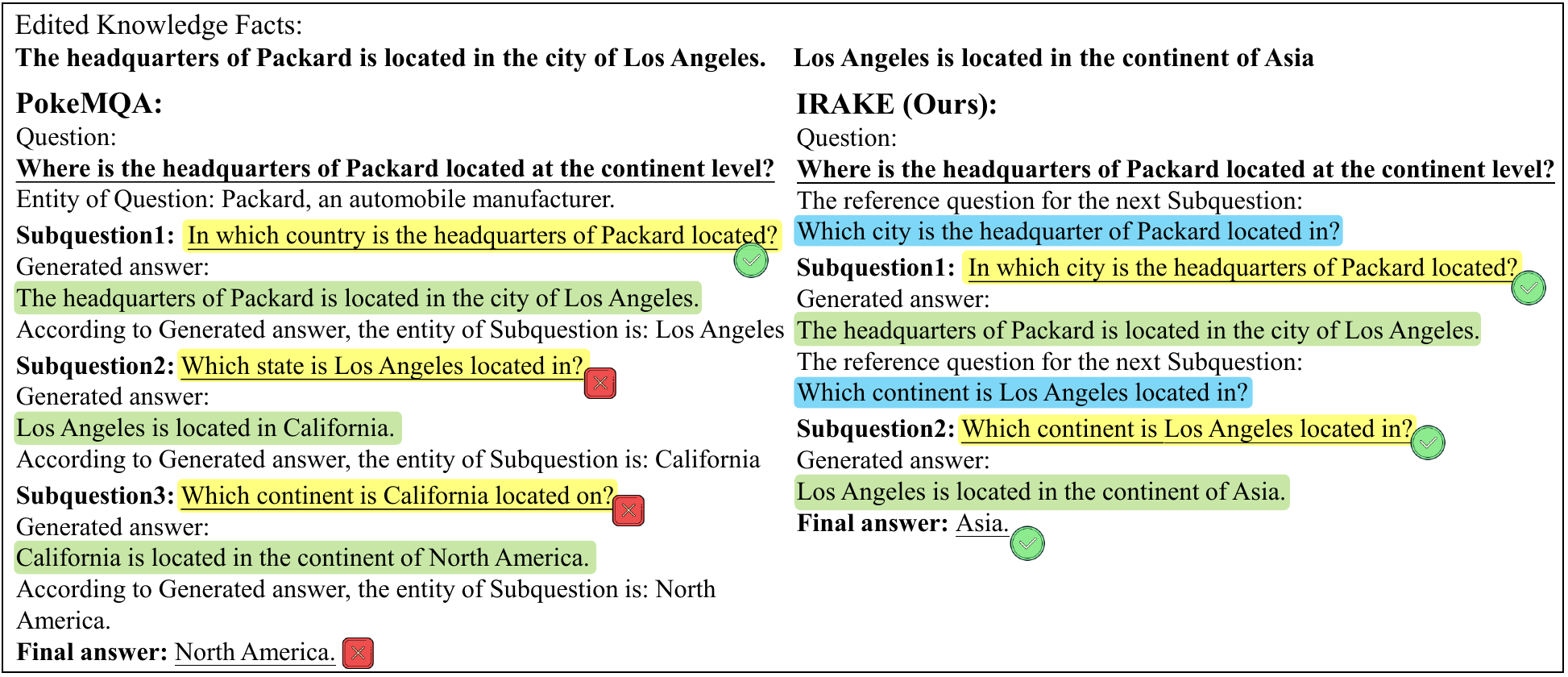}
\caption{Case study of PokeMQA and \modelname to solve a $2$-hop question in MQUAKE-2002. Yellow texts are the decomposed subquestions. Green texts are the answers generated by the LLM or retrieved from the edited fact memory, and blue texts are the reference subquestions used for decomposition guidance from the edited fact level.}
\label{fig:case_study}
\end{figure*}

\subsection{Further Discussions}
\label{sec:further_dis}
\paragraph{Can the decomposition guidance in \modelname truly help LLMs overcome the ``edit skipping'' issue, thereby improving the performance of LLMs on MQuAKE?}
To investigate this issue, we define a metric called the recall of edited facts (abbr. Recall), which represents the proportion of all facts required to be edited for a complex question that are successfully retrieved by the model during the reasoning process.
We compare \modelname, \modelname without case guidance (denoted by ``$-$\,CaseG''), \modelname without fact guidance (denoted by ``$-$\,FactG''), and PokeMQA on the two datasets, as shown in Figure~\ref{fig:recall_acc_exp}.
The results show a positive correlation between Recall and Accuracy, suggesting that retrieving more correct edited facts during reasoning indeed leads to better performance.
Moreover, the proposed fact- and case-level guidance both help LLMs more accurately hit the facts to be edited during reasoning, effectively mitigating the ``edit skipping'' issue.

\paragraph{We analyze the impact of the similarity threshold $\theta$ used to select similar edited cases for guidance.}
As shown in Figure~\ref{fig:exp3}, the X-axis denotes different threshold values $\theta$. The bar chart shows the number of test cases guided by similar cases, while the line chart depicts model performance (Acc).
As the similarity threshold $\theta$ decreases, more test cases are guided by similar edited cases.
As the number of decompositions guided by similar cases increases, the model's performance shows a certain degree of improvement. As the $\theta$ decreases, the number of cases in the test set that are guided increases significantly, yet the performance does not show a corresponding substantial improvement. This is primarily because cases with lower similarity provide limited guidance.

\paragraph{Performance analysis of the KE methods in the $n$-hop question answering task.}
We evaluate the performance of \modelname and three strong baselines on questions with varying hop numbers under the all-edited setting, as shown in Figure~\ref{fig:multi_hop_exp}.
Our method consistently achieves superior performance across questions with varying hops, particularly on more complex $4$-hop cases, which we attribute to its guided decomposition that mitigates error propagation from intermediate subquestions.

\paragraph{Additional analysis and results.}
We also conduct several other discussions: the effect of varying the number of training examples used to construct the edited case library (see Appendix~\ref{app:exp_per_analysis_the_edited_case_library});
the advantage of storing questions over edited factual answers for edited facts guidance (see Appendix~\ref{app:exp_guidance_strategies_edited_facts});
the effectiveness of similarity-based selection compared to other retrieval strategies used for edited case guidance (see Appendix~\ref{app:exp_effect_similarity_selection});
the impact of different base LLMs on our model's performance (see Appendix~\ref{app:exp_performance_analysis_different_baseLLMs});
the effectiveness of fact-guided decomposition on middle-hop edits (see Appendix~\ref{app:effectiveness_middle_hop});
and the analysis of fact-guided decomposition module across different numbers of pre-retrieved facts (see Appendix~\ref{app:exp_num_of_pre}).

\subsection{Case Study}
We conduct a case study as presented in Figure~\ref{fig:case_study}, where the input question involves two edited facts. Both \modelname and PokeMQA successfully generate the first-step subquestion. However, the first-step subquestion decomposed by PokeMQA inquires about ``which country''. Although the correct edited fact is retrieved, the subquestion does not fully match the retrieval result. In contrast, \modelname retrieves the correct edited fact in pre-retrieval and utilizes its corresponding atomic question to guide the subquestion decomposition. As a result, the generated subquestion better aligns with the subsequently retrieved edit.

During the decomposition of the second subquestion, PokeMQA generates a subquestion: ``Which state is Los Angeles located in?'' with finer granularity than the edited fact: ``Los Angeles is located in the continent of Asia''.
This mismatch causes the LLM to skip the edit, resulting in an incorrect final answer.
In contrast, \modelname retrieves the correct edited fact during the second step of pre-retrieval and utilizes its corresponding atomic question to decompose the subquestion that aligns with the edited fact, ultimately leading to the correct answer.

\section{Conclusion}
\label{sect:conclusion}
In this paper, we introduce the challenge of edit skipping caused by the knowledge granularity mismatch in MQuAKE. We propose an iterative retrieval-augmented KE method based on edit-guided question decomposition from the perspectives of edited facts and cases.
We also design a state backtracking mechanism to alleviate the impact of failed guidance. Experimental results demonstrate that our method alleviates the ``edit skipping'' issue in MQuAKE and outperforms state-of-the-art baselines.

\section*{Acknowledgments}
This work was funded by National Natural Science Foundation of China (No. 62272219).

\section*{Limitations}
\label{sect:limit}
Our work has the following limitations:
First, \modelname focuses on improving KE of multi-hop question answering by enhancing query decomposition with the fact-level and case-level guidance. However, it does not directly enhance the ability of LLMs to decompose problems or answer questions based on context. The effectiveness of KE is also influenced by the underlying LLM, as reflected in Section~\ref{sect:main_res}.
Second, for the case-level guidance, \modelname builds an edited case library by storing historical decomposition records during the entire editing process. We initialize the edited case library by sampling a small set of cases from the training set. We will investigate how to mitigate the cold start problem in the absence of high-quality training data in future work.

\bibliography{custom}

\appendix


\section{Dataset Statistics}
\label{app:Dataset Statistics}
The detailed statistics of the datasets in our experiments are shown in Table~\ref{tab:detailed_dataset_dist}.
We do not use the MQuAKE-3k dataset due to its high number of conflicts, which makes it difficult to accurately assess the performance of KE methods (see Table~\ref{tab:dataset_stats}). MQuAKE-2002~\citep{wang24deepedit}, by removing conflicting instances, provides a reliable evaluation of performance, while MQuAKE-hard evaluates the capability to handle complex cases.

\begin{table}[!ht]
\centering
\resizebox{\columnwidth}{!}{
\begin{tabular}{l|ccccc}
\toprule
Datasets & \#Edits & 2-hop & 3-hop & 4-hop & Total \\
\midrule
\multirow{5}{*}{MQuAKE-2002} & 1 & 479 & \ \ 71 & \ \ \ \ 7 & \ \ 557 \\
& 2 & 487 & 244 & \ \ 20 & \ \ 751 \\
& 3 & - & 310 & 116 & \ \ 426 \\
& 4 & - & - & 268 & \ \ 268 \\
& All & 966 & 625 & 411 & 2,002 \\
\midrule
MQuAKE-hard & 4 & - & - & 429 & \ \ 429 \\
\bottomrule
\end{tabular}
}
\caption{Statistics of the datasets in the experiments.}
\label{tab:detailed_dataset_dist}
\end{table}

\begin{table}[!ht]
\centering
\resizebox{\columnwidth}{!}{
\begin{tabular}{l|rccc}
\toprule
Datasets & \#Inst. & \makecell{Avg.\\hops} & \makecell{Avg.\\edits} & \#Conflicts \\
\midrule
MQuAKE-3k & 3,000 & 3.0 & 2.0 & 998 \\
MQuAKE-2002 & 2,002 & 2.7 & 2.2 & \ \ \ \ 0 \\
MQuAKE-hard & 429 & 4.0 & 4.0 & \ \ \ \ 0 \\
\bottomrule
\end{tabular}
}
\caption{Difference between the datasets.}
\label{tab:dataset_stats}
\end{table}

\begin{table}[!t]
\centering
\resizebox{\columnwidth}{!}{
\begin{tabular}{l|cc|cc}
\toprule
\text{Datasets} & \multicolumn{2}{c|}{\text{MQuAKE-2002}} & \multicolumn{2}{c}{\text{MQuAKE-hard}} \\
\cmidrule(lr){1-1} \cmidrule(lr){2-3} \cmidrule(lr){4-5}
  Methods            & \makecell{Runtime \\ (Avg. case)} & Acc & \makecell{Runtime \\ (Avg. case)} & Acc  \\
\midrule
MeLLo             & \ \ 7.2 (s) & 14.33  & \ \ 8.5 (s)  & 4.60   \\
DeepEdit          & 18.1 (s) & 17.76  & 18.9 (s)  & 8.80   \\
PokeMQA           & \ \ 5.4 (s)  & 36.81  & \ \ 6.6 (s) & 27.16  \\
\modelname (Ours) & 14.6 (s) & 55.24 & 18.3 (s) & 40.79  \\
\bottomrule
\end{tabular}
}
\caption{Analysis of runtime for different models.}
\label{tab:analysis_of_runtime}
\end{table}

\begin{table}[!t]
\centering
\resizebox{\columnwidth}{!}{
\begin{tabular}{l|c|c}
\toprule
\text{Methods} & \text{Input (Inference)} & \text{Output (Avg. case)} \\
\midrule
MeLLo             & \ \ \,756 (tokens)   & 172 (tokens)    \\
DeepEdit          & 1,254 (tokens)  & 108 (tokens)    \\
PokeMQA           & \ \ \,589 (tokens)   & \ \ 92 (tokens)     \\
\modelname (Ours) & 1,859 (tokens)  & \ \ 84 (tokens)     \\
\bottomrule
\end{tabular}
}
\caption{Number of tokens in inference and average number of output tokens for each test edited case.}
\label{tab:analysis_of_token}
\end{table}

\begin{table*}[t]
\centering
\resizebox{\textwidth}{!}{
\begin{tabular}{l|ll|ll|r}
\toprule
\text{Datasets} & \multicolumn{2}{c|}{\text{MQuAKE-2002}} & \multicolumn{2}{c|}{\text{MQuAKE-hard}} \\
\cmidrule(r){1-1} \cmidrule(r){2-3} \cmidrule(lr){4-5} \cmidrule(lr){6-6}
\#Used / \#All          & Acc    & Hop-Acc& Acc    & Hop-Acc & Memory Usage   \\
\midrule
\ \ \ \ 0 / 9218   (0.0$\%$)      & 51.59  & 37.95  & 37.06  & 32.40  & 0 (tokens)    \\
100 / 9,218 (1.1$\%$)      & 53.84 (4.36$\%\uparrow$)  & 41.75 (10.01$\%\uparrow$)  & 37.99 (2.51$\%\uparrow$)  & 33.10 (2.16$\%\uparrow$)  & 9,605 (tokens) \\
300 / 9,218 (3.3$\%$)       & 54.59 (5.82$\%\uparrow$)	  & 43.20 (13.83$\%\uparrow$)  & 39.16 (5.67$\%\uparrow$)  & 34.49 (6.45$\%\uparrow$)  & 28,970 (tokens) \\
500 / 9,218 (5.4$\%$)       & 55.24 (7.08$\%\uparrow$)  & 44.80 (18.05$\%\uparrow$)  & 40.79 (10.06$\%\uparrow$)  & 35.90 (10.8$\%\uparrow$)  & 46,036 (tokens) \\
700 / 9,218 (7.6$\%$)       & 56.74 (9.98$\%\uparrow$)  & 46.40 (22.27$\%\uparrow$)  & 43.12 (16.35$\%\uparrow$)  & 38.22 (17.96$\%\uparrow$)	  & 64,339 (tokens) \\
\bottomrule
\end{tabular}
}
\caption{Performance analysis of \modelname w.r.t. the number of training cases used  (\# denotes ``number'') and the memory usage in edited case records.}
\label{tab:performance_different_number_of_training_cases}
\end{table*}

\section{Implementation Details}
\label{app:inple-details}
This section provides the implementation details.
For \modelname and all baselines, 
we use LLaMa-3-8B-instruct~\citep{dubey24llama3}, DeepSeek-V2-Lite-16B~\citep{deepseek24deepseek}, and GPT-4o-Mini~\citep{hurst24gpt4o} as the base LLM alternatively for a fair and comprehensive comparison.
The former two are among the most popular open-source LLMs and the latter one is a popular and cost-efficient black-box LLM.
Since we do not introduce any innovations in the retrieval model, \modelname utilizes the models and methods from previous work for retrieval.
Specifically, during the pre-retrieval phase, \modelname directly employs the encoder provided by PokeMQA~\citep{gu24pokemqa} to calculate the similarity between question $Q$ and the edited facts in the edited fact memory.
The top 3 most similar pieces of facts are selected for subsequent judgment in this phase.
In the fine-grained retrieval phase, \modelname adopts the retrieval method from PokeMQA (more details can be found in the relevant paper). 
For the retrieval of similar edited cases, which involves calculating the similarity between questions, we directly use the general model, mxbai-embed-large-v1~\citep{li2023angle}, to encode the questions and compute their similarity, where the threshold $\theta$ is set to 0.80. 
To build the edited case library, we randomly sample 500 cases from the training set.
The hyperparameters of \modelname during inference are configured as follows: 
temperature is set to 0, max-tokens is set to 200, and repetition-penalty is set to 1.1.

\section{Prompts of \modelname}
\label{app:prompt}
\begin{figure*}
\centering
\includegraphics[width=\linewidth]{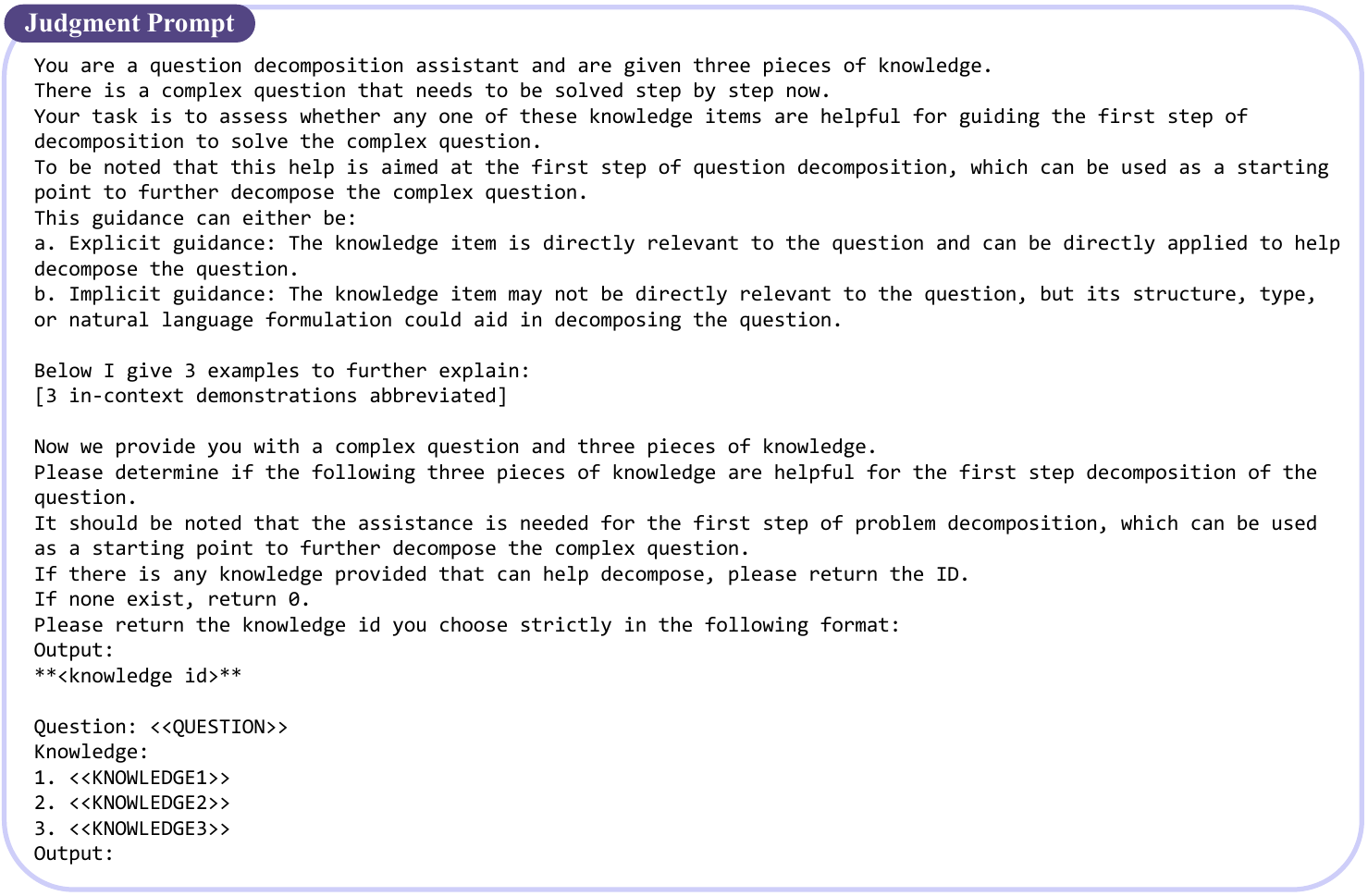}
\caption{Prompt for judging helpful knowledge used for question decomposition.}
\label{fig:prompt-judge}
\end{figure*}

\begin{figure*}
\centering
\includegraphics[width=\linewidth]{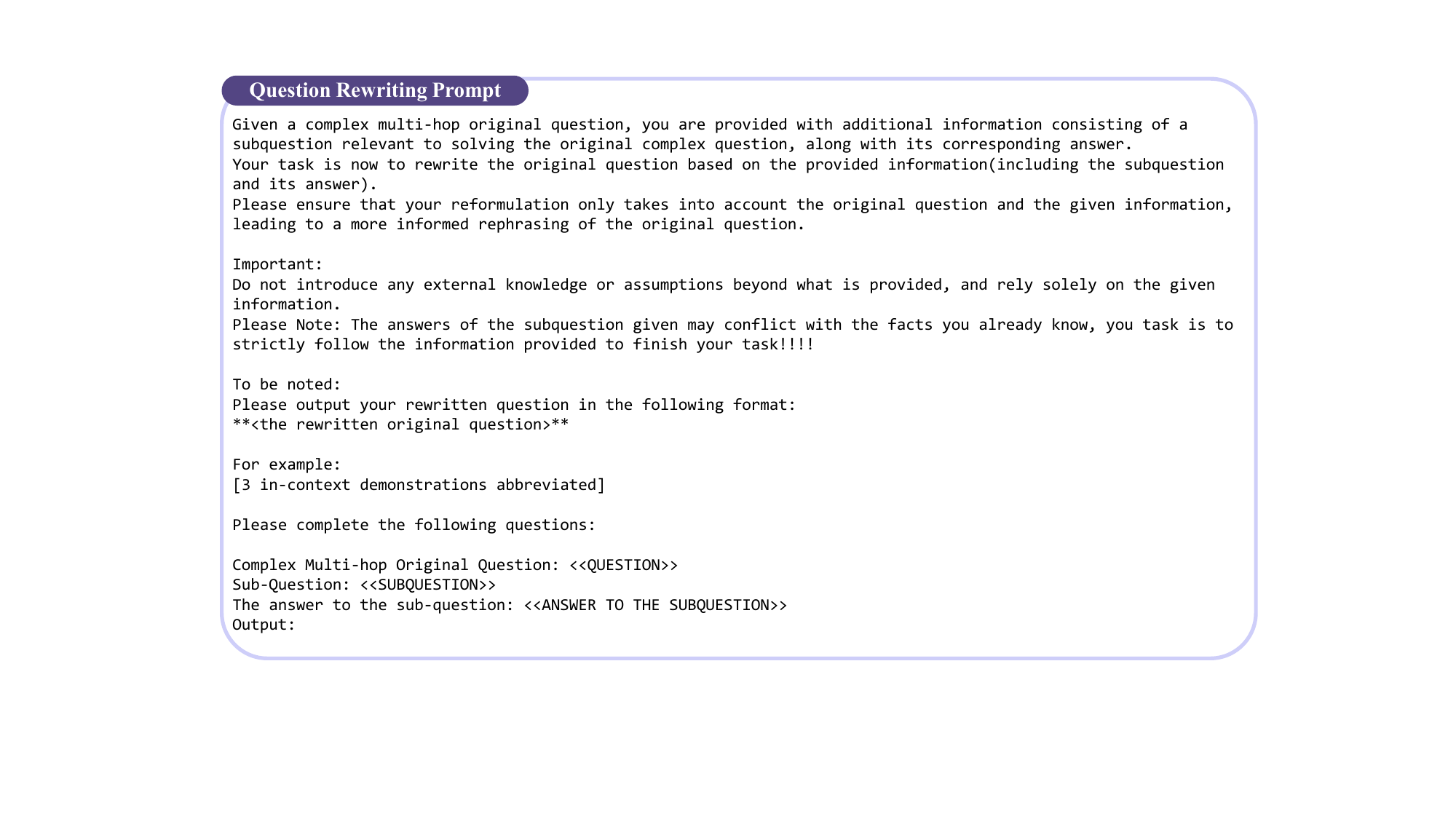}
\caption{Prompt for question rewriting.}
\label{fig:prompt-rewrite}
\end{figure*}

\begin{figure*}
\centering
\includegraphics[width=\linewidth]{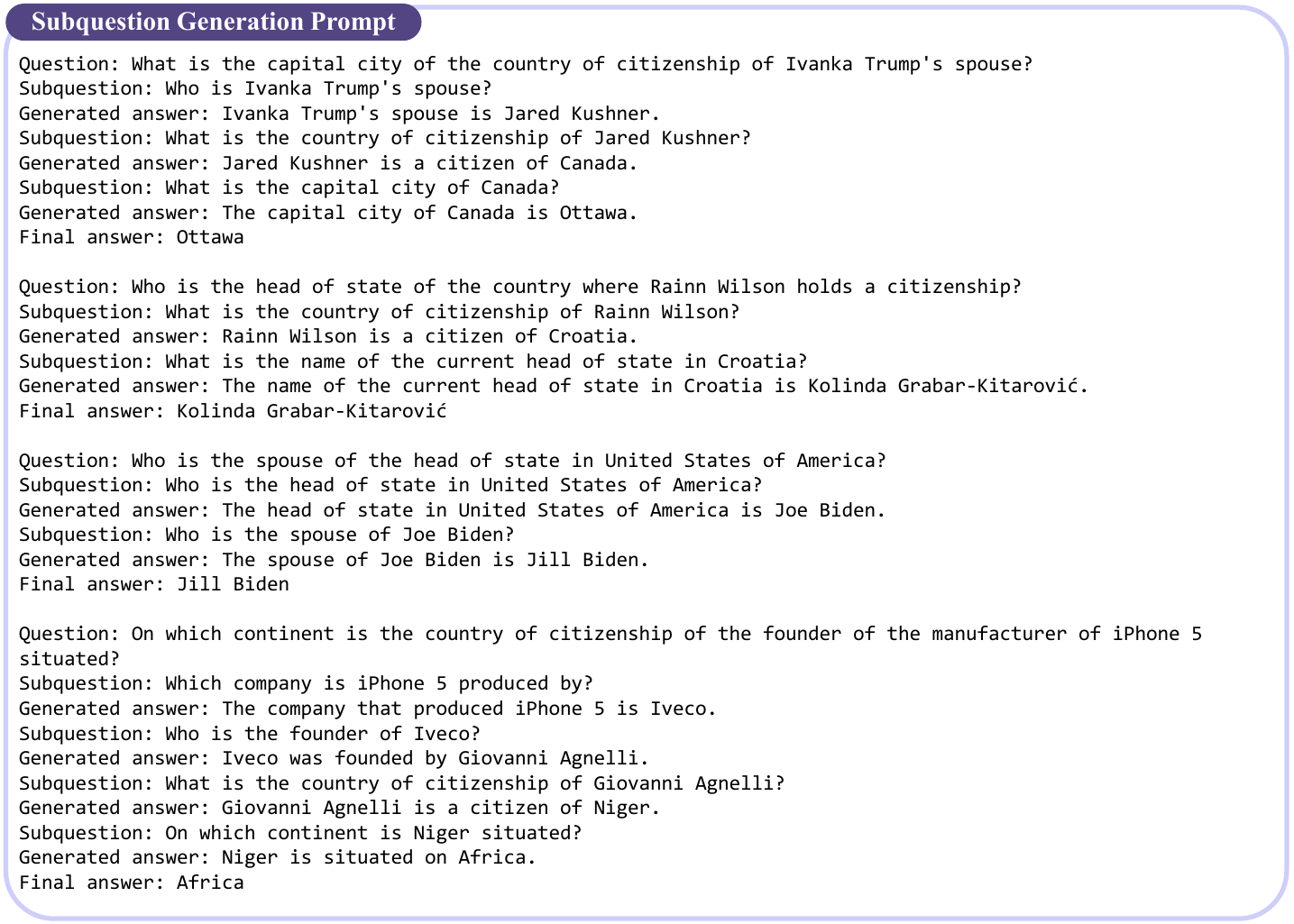}
\caption{Prompt for question decomposition without guidance.}
\label{fig:prompt-static}
\end{figure*}

\begin{figure*}
\centering
\includegraphics[width=\linewidth]{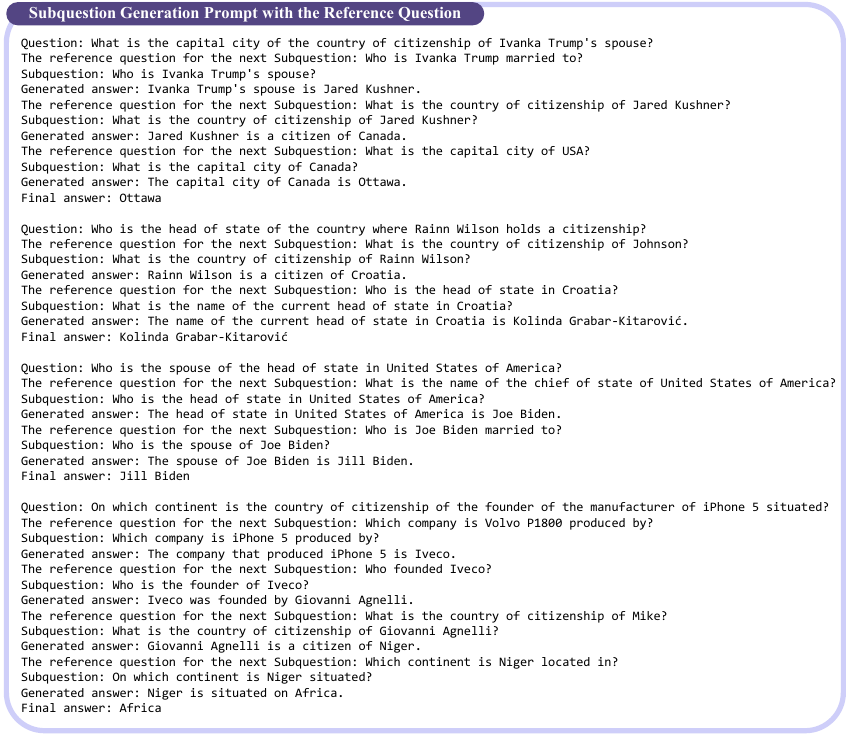}
\caption{Prompt for question decomposition with guidance from the edited fact level.}
\label{fig:prompt-reference}
\end{figure*}

Figure~\ref{fig:prompt-judge} provides the prompt for LLM to evaluate and identify the most relevant edit fact for edited fact-level guidance.
Figure~\ref{fig:prompt-rewrite} provides the prompt for LLM to rewrite the original question given the subquestion and its corresponding answer for the next pre-retrieval.
Figure~\ref{fig:prompt-static} is the prompt for question decomposition without guidance. 
Figure~\ref{fig:prompt-reference} is the prompt for question decomposition with guidance from the edited fact level, which uses the atomic question corresponding to the edited fact for guidance. 
 
\section{Details About Experiments}
\label{app:expdetals}
The parameter updating KE methods in our experiments, including FT, ROME, and MEMIT, are all implemented with the EasyEdit library~\citep{wang2023easyedit}.
We follow the default hyperparameter settings on LLaMa-3-8B in the library.
The inference hyperparameters of the retrieval-augmented KE methods in our experiments are the same as those of our \modelname for a fair comparison.

\section{Runtime and Token-Level Comparison Across Models}
\label{app:exp_runtime}
To provide an evaluation of the computational efficiency of our method, we present a comparison of runtime and token-level statistics across models. 
Specifically, we analyze the runtime cost and the number of input/output tokens involved in each edited case. 
Since \modelname is built upon an iterative retrieval-augmented framework, we compare \modelname with baseline methods of the same type without loss of generality.

The detailed runtime results are shown in Table~\ref{tab:analysis_of_runtime}. 
Due to the additional pre-retrieval and question rewriting operations performed in each iteration, IRAKE requires longer inference time compared to the methods such as MeLLo and PokeMQA. 
Due to the more streamlined process, \modelname requires less inference time than DeepEdit.

The token-level comparison results are shown in Table~\ref{tab:analysis_of_token}. We report the average number of input tokens during inference in each iteration, as well as the average number of output tokens per edited case.
Token counts are computed using the tiktoken library, with the encoding model uniformly set to gpt-3.5-turbo. 
Similarly, due to the additional operations performed in each iteration, \modelname requires more input tokens during inference compared to other methods. 
However, thanks to its efficient workflow, \modelname can generate fewer output tokens to arrive at the final answer.

To address the issue of edit skipping, we argue that the most direct approach is to provide context related to edits. 
\modelname tackles this problem by employing edited facts guidance (e.g., pre-retrieval and question rewriting) and similar edited case records guidance, which indeed results in longer runtime and increased consumption of tokens. 
Moreover, the number of tokens stored in the edited knowledge fact memory is 23,540 for the baseline, compared to 44,539 for our \modelname. 
However, \modelname generates fewer output tokens per edited case on average. 
Considering the performance improvements, we believe these additional costs are still worthwhile.

\section{Performance Analysis About the Number of Training Cases for the Edited Case Library}
\label{app:exp_per_analysis_the_edited_case_library}
To better understand the impact of the edited case library size on model performance, we conduct a detailed analysis using different numbers of training examples to construct the library. 
Specifically, we sample the subsets of varying sizes from a training set of 9,218 examples and evaluate the resulting performance of our model, \modelname.

The results are presented in Table~\ref{tab:performance_different_number_of_training_cases}. 
As the number of training cases used to build the case library increases, the performance of \modelname improves accordingly. 
Notably, even with a relatively small number of examples, \modelname achieves significant performance gains, demonstrating its ability to effectively leverage limited edited cases. 
This analysis highlights the efficiency and scalability of our case-based guidance design.

\section{Impact of Different Guidance Strategies Using Edited Facts}
\label{app:exp_guidance_strategies_edited_facts}
In this section, we analyze the impact of different guidance strategies when leveraging edited facts for guidance.
The experimental results are shown in Table~\ref{tab:performance_different_guidance_ways}. 
``w/o guided'' represents the scenario without guidance, ``w/ factual answer'' represents guidance using edited factual answers, and ``w/ question'' represents the method proposed in our paper, which uses a question according to the edited fact as guidance.
The edited factual answers often appear counterfactual or unnatural to LLMs, potentially leading to confusion during multi-hop reasoning. 
Question-based guidance provides a more natural form of supervision for decomposition. 
The experimental results show that while using edited factual answers offers some improvement over no guidance, it performs worse than the question-based guidance. 
This supports our design choice of guiding decomposition using stored questions rather than factual answers.

\begin{table}[!t]
\centering
\resizebox{\columnwidth}{!}{
\begin{tabular}{l|cc|cc}
\toprule
\text{Datasets} & \multicolumn{2}{c|}{\text{MQuAKE-2002}} & \multicolumn{2}{c}{\text{MQuAKE-hard}} \\
\cmidrule(lr){1-1} \cmidrule(lr){2-3} \cmidrule(lr){4-5}
Method: \modelname            & Acc & Hop-Acc & Acc & Hop-Acc  \\
\midrule
w/o guided             & 42.10  & 31.46  & 33.96  & 30.70   \\
w/ factual answer  & 50.20  & 36.96  & 37.53  & 32.63   \\
w/ question    & \textbf{55.24}  & \textbf{44.80}  & \textbf{40.79}  & \textbf{35.90} \\
\bottomrule
\end{tabular}
}
\caption{Performance comparison of different guidance strategies.}
\label{tab:performance_different_guidance_ways}
\end{table}

\begin{table}[!h]
\centering
\resizebox{\columnwidth}{!}{
\begin{tabular}{l|cc|cc}
\toprule
\text{Datasets} & \multicolumn{2}{c|}{\text{MQuAKE-2002}} & \multicolumn{2}{c}{\text{MQuAKE-hard}} \\
\cmidrule(lr){1-1} \cmidrule(lr){2-3} \cmidrule(lr){4-5}
Method: \modelname            & Acc & Hop-Acc & Acc & Hop-Acc  \\
\midrule
w/o edited case           & 51.59  & 37.95  & 37.06  & 32.40   \\
w/ random edited case     & 52.60  & 41.26  & 38.92  & 34.03   \\
w/ similar edited case    & \textbf{55.24} & \textbf{44.80} & \textbf{40.79} & \textbf{35.90} \\
\bottomrule
\end{tabular}
}
\caption{Performance analysis of \modelname with different ways to get edited cases for guidance.}
\label{tab:performance_different_type_cases}
\end{table}

\section{Effectiveness of Similarity-Based Edited Case Selection}
\label{app:exp_effect_similarity_selection}
Existing works~\citep{Bangzhengapp, Junjieapp} suggest that reasoning for complex questions can be expressed through inference paths that incorporate triple-based knowledge, exhibiting a certain structure of question decomposition, with the semantic information of the question embedded in the path of triples.
Consequently, semantically similar complex questions are more likely to share similar reasoning paths that can be edited and the potential edited points for edited facts tend to overlap (as mentioned in Section~\ref{sect:dynamic_prompt}).

We conduct experiments to validate our claims, with the results shown in Table~\ref{tab:performance_different_type_cases}. 
All methods are given the same number of edited cases for guidance. 
``w/o edited case'' denotes the model without any guidance. 
``w/ random edited case'' randomly replaces the most similar case with another edited case, while ``w/ similar edited case'' uses the most similar edited case for guidance. 
The experimental results show that even without using the most similar edited case for guidance, the model's performance still improves to some extent, due to the increase in the number of demonstrations in in-context learning. 
However, using similar edited cases for guidance yields the best results.

\begin{table}[!t]
\centering
\resizebox{\columnwidth}{!}{
\begin{tabular}{l|ccc|ccc}
\toprule
\text{Datasets} & \multicolumn{3}{c|}{\text{MQuAKE-2002}} & \multicolumn{3}{c}{\text{MQuAKE-hard}} \\
\cmidrule(r){1-1} \cmidrule(r){2-4} \cmidrule(lr){5-7}
Base LLMs         & Recall & \#Doubts & Acc & Recall & \#Doubts & Acc \\
\midrule
LLaMa-3-8B   & 55.83  & 279  & 55.24 & 44.23 & 169 & 40.79  \\
DeepSeek-V2  & 59.60  & 312  & 50.51 & 49.79 & 183 & 40.09  \\
GPT-4o-Mini           & 62.64  & 335  & 46.75 & 53.87 & 194 & 32.13  \\
\bottomrule
\end{tabular}
}
\caption{Analysis of \modelname under different base LLMs.}
\label{tab:performance_different_baseLLMs}
\end{table}

\section{Performance Analysis Across Different Base LLMs}
\label{app:exp_performance_analysis_different_baseLLMs}
We conduct additional experiments to analyze how different base LLMs affect the performance of our model. 
The results are summarized in Table~\ref {tab:performance_different_baseLLMs}.

We first analyze the impact of different LLMs on the successful judgment and selection of edited facts for guidance during the pre-retrieval step. 
We use the recall of edited facts as the evaluation metric. 
This metric represents the proportion of all facts required to be edited for a complex question that are successfully selected by LLMs during the pre-retrieval step.
Overall, as the base LLMs' reasoning ability improves, recall also increases, which is consistent with expectations.

\begin{table}[!t]
\centering
\resizebox{\columnwidth}{!}{
\begin{tabular}{l|cc|cc}
\toprule
\text{Datasets} & \multicolumn{2}{c|}{\makecell{\text{MQuAKE-2002}\\(\#: 411)}} & \multicolumn{2}{c}{\makecell{\text{MQuAKE-hard}\\(\#: 429)}} \\
\cmidrule(lr){1-1} \cmidrule(lr){2-3} \cmidrule(lr){4-5}
Method: \modelname       & Recall & Acc    & Recall & Acc  \\
\midrule
w/o fact-guided           & 47.64  & 33.09  & 51.44  & 33.96   \\
w/ fact-guided         & \textbf{56.71}  & \textbf{43.79} & \textbf{56.23} & \textbf{40.79}  \\
\bottomrule
\end{tabular}
}
\caption{Effectiveness of fact-guided decomposition on multi-hop questions with middle-hop edited facts.}
\label{tab:effectiveness_middle_hop}
\end{table}

However, a higher recall does not always translate into better overall performance.
The reasons may be complex, but through analysis, we find that one possible reason could be the varying degrees of sensitivity to counterfactuals across different base LLMs. 
Some models may treat edited facts with excessive skepticism, which can hinder their ability to generate appropriate subquestions. 
For instance, given the question ``What is the capital of the country where American Ninja Warrior originated?'', the edited fact ``The capital of the United Kingdom is Angri.'' may trigger unnecessary doubt. 
A counterfact-sensitive model might respond with a subquestion like ``Is Angri the correct capital of the United Kingdom?'', disrupting the intended reasoning path.

To explore this, we conduct a simple statistical analysis of some common forms of doubt expressions, such as ``(Note: ***)'' or ``Is *** correct?'', and count their occurrences in model outputs (\#Doubts). 
In Table~\ref {tab:performance_different_baseLLMs}, the models that exhibit more such expressions tend to perform worse, suggesting that increased sensitivity to counterfactuals may negatively impact reasoning effectiveness.

\section{Effectiveness of Fact-Guided Decomposition on Middle-Hop Edits}
\label{app:effectiveness_middle_hop}

We evaluate the effectiveness of fact-guided decomposition on middle-hop edits.
We first identify the number of cases in which edited facts occur in a middle hop across different datasets. Specifically, there are 411 such 4-hop cases in MQuAKE-2002, and 429 such 4-hop cases in MQuAKE-hard. 
We then evaluate model performance with and without fact-guided decomposition on these cases, reporting both final accuracy (Acc) and the recall of edited facts (Recall), as defined in Section~\ref{sec:further_dis}. Recall measures the proportion of all required edited facts for a given multi-hop question that are successfully captured during reasoning. All experiments are conducted using LLaMa3-8B-Instruct. 
The results are shown in Table~\ref{tab:effectiveness_middle_hop}. We observe that applying fact-guided decomposition significantly improves both the recall of edited facts and the final accuracy, demonstrating its utility in handling middle-hop edits.

\section{Analysis of Fact-Guided Decomposition Module Across Different Numbers of Pre-retrieved Facts}
\label{app:exp_num_of_pre}

\begin{table}[!t]
\centering
\resizebox{\columnwidth}{!}{
\begin{tabular}{l|c|c|c}
\toprule
\makecell{Top-$k$\\ pre-retrieved facts} & \makecell{Recall in\\ pre-retrieval} & \makecell{Acc in\\ judgment} & Acc  \\
\midrule
$k = 1$                     & 46.69  & \textbf{86.32}  & 51.09 \\
$k = 3$ (default)           & 58.17  & 83.28  & \textbf{55.24} \\
$k = 5$                     & \textbf{67.47}  & 75.89  & 54.02 \\
\bottomrule
\end{tabular}
}
\caption{Analysis of pre-retrieval, judgment in the fact-guided decomposition module, and final accuracy under different top-$k$ settings.}
\label{tab:num_of_pre}
\end{table}

We conduct an analysis to better understand the fact-guided decomposition module by varying the number of pre-retrieved facts. Specifically, we compare the performance of the module under different top-$k$ pre-retrieval settings ($k = 1, 3, 5$) on the MQuAKE-2002 dataset using LLaMa3-8B-Instruct. For each setting, we report three metrics:
\begin{itemize}
    \item \textbf{Recall in pre-retrieval}: the recall of required edited facts retrieved in the pre-retrieval step;

    \item \textbf{Acc in judgment}: whether the model accurately determines the correct guidance behavior. If the required edited fact is not retrieved in the pre-retrieval step, the model should output 0 (no guidance); otherwise, it should correctly select the required edited fact;

    \item \textbf{Acc}: the overall performance of the model.
\end{itemize}

As shown in Table~\ref{tab:num_of_pre}, increasing $k$ improves the recall of edited facts during the pre-retrieval step, but also makes it more challenging for the model to make correct guidance decisions. To achieve better overall performance, the choice of $k$ should be made with a trade-off between recall and judgment accuracy.

\end{document}